\documentclass[conference]{IEEEtran} 

\IEEEoverridecommandlockouts              
\usepackage{xcolor}
\long\def\invis#1{}

\usepackage{amssymb}
\usepackage{mathtools}
\usepackage{graphicx}
\usepackage{float}
\usepackage{tikz}
\usepackage{listings}
\usepackage{hyperref}
\usepackage{graphicx}
\usepackage{cite}
\usepackage{dsfont}
\usepackage{mathtools,etoolbox}
\usepackage[ruled, linesnumbered]{algorithm2e}
\usepackage[none]{hyphenat}
\usepackage[utf8]{inputenc}
\usepackage[english]{babel}
\usepackage[T1]{fontenc}
\usepackage{mathtools, nccmath}

\usepackage{amsthm}
\usepackage{xfrac}
\usepackage{nicefrac}
\usepackage{booktabs}
\usepackage[switch, pagewise]{lineno}
\usepackage{mathrsfs}
\DeclareMathAlphabet{\mathpzc}{OT1}{pzc}{m}{it}
\usepackage{multirow}
\usepackage{multicol}
\usepackage{bbm}
\usepackage{soul}
\usepackage{xfrac}
\usepackage{balance}
\usepackage{titlesec}
\usepackage[final]{pdfpages}
\usepackage{amsmath}
\usepackage{xcolor}

\usepackage{cite}
\usepackage{amsmath,amssymb,amsfonts}
\usepackage{algorithmic}
\usepackage{graphicx}
\usepackage{mathtools}
\usepackage{textcomp}
\usepackage{xcolor}
\usepackage{url}
\usepackage[affil-it]{authblk}

\def\BibTeX{{\rm B\kern-.05em{\sc i\kern-.025em b}\kern-.08em
    T\kern-.1667em\lower.7ex\hbox{E}\kern-.125emX}}

\begin{document}

\title{uScenes: A Multimodal RGB and 3D Sonar Dataset for Underwater Robot Perception
\thanks{$^{1}$Embodied Robotics and Automation Lab, University of South Florida, Tampa, FL 33610, USA. Emails: \texttt{\{dongt, xlin2\}@usf.edu}.}
\thanks{$^{2}$Mechanical and Aerospace Engineering Department, University of Florida, Gainesville, FL 32611, USA}
\thanks{$^{3}$Naval Architecture and Ocean Engineering at the Seoul National University, Seoul 08826, South Korea 
{\tt \{janeshin\}@snu.ac.kr}}
}
\author{Trung Tien Dong\textsuperscript{1}, Zhenqi Wu\textsuperscript{1}, Aditya Penumarti\textsuperscript{2}, Zi-Hao Zhang\textsuperscript{2}, Micaiah Bartlett\textsuperscript{1},  
Jane Shin\textsuperscript{3}, Xiaomin Lin\textsuperscript{1}
}


\maketitle
\thispagestyle{empty}
\pagestyle{empty}

\begin{abstract}

Robust perception is essential for the deployment of autonomous underwater robots. However, optical cameras become unreliable under poor illumination and backscatter. Forward looking (2D) acoustic sensors remain effective under these conditions, but they measure range and bearing while leaving elevation unresolved, creating an ambiguity that prevents individual sonar returns from being localized in three dimensional (3D) space. This complicates the sensor use for 3D scene understanding and precise object detection. We introduce \textbf{uScenes}, a multimodal underwater dataset containing synchronized 3D multibeam sonar point clouds and RGB imagery. The dataset contains 110 scenes and 95,834 synchronized observation, representing 277.6 minutes of data collected across multiple field sessions. uScenes establishes a foundation for underwater sensor fusion, cross modal representation learning and 3D scene understanding. Code and
datasets are given at \url{https://github.com/era-research-lab/uScenes}.

\end{abstract}

\section{Introduction}
\label{sec:introduction}


Underwater robots are increasingly used to perform missions that are difficult, costly, or dangerous for human divers, including marine infrastructure inspection~\cite{Huang2025BrSPCD}, environmental monitoring~\cite{lin2022oystersim}, habitat observation~\cite{sadrfaridpour2021detecting}, mapping~\cite{burgul2025compact3dsonar}, search and recovery~\cite{cao2016multiauv,simetti2014trident}, and navigation~\cite{karapetyan2021human}. The success of these missions depends on the ability of a robot to interpret its surroundings despite poor illumination, suspended particles, changing water conditions, and complex environments. Under such conditions, optical cameras are not as helpful, even though they provide rich appearance, texture, and semantic information. This is because light attenuation, color distortion, and backscatter can substantially degrade visual observations~\cite{Hao2024UnderwaterOptical}. A robot that relies primarily on optical sensing may therefore lose important perceptual information when environmental conditions become variable or adverse.

\begin{figure} 
    \centering 
    \includegraphics[width=1\columnwidth]{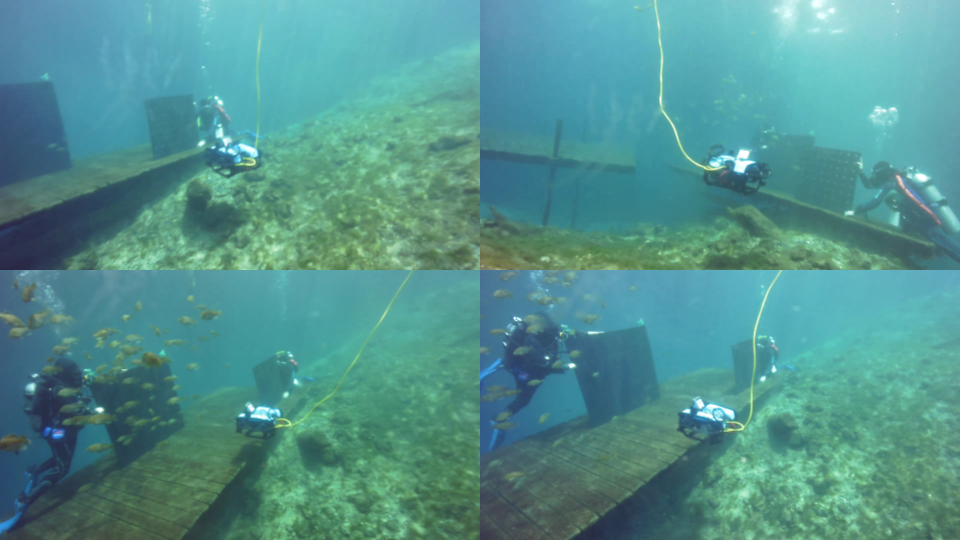} 
    \caption{Field collection of the dataset at Blue Grotto, Florida.} 
    \vspace{-5mm} 
    \label{fig:illustrative_example} 
\end{figure}

Acoustic sensors remain effective under many of these conditions and are widely used for underwater perception~\cite{Huy2023UnderwaterPerception}. However, forward looking sonar (2D imaging) sonars measure range and bearing while leaving elevation unresolved. An individual return can therefore correspond to multiple possible positions within the vertical field of view rather than a unique location in three dimensional (3D) space~\cite{Huang2015ASFM}. This ambiguity complicates the use of conventional sonar measurements for 3D scene understanding and precise spatial localization of detected objects.

3D multibeam imaging sonar provides a different acoustic representation \cite{burgul2025compact3dsonar}. By measuring returns across both horizontal and vertical angular dimensions, it allows each valid range observation to be mapped directly to a 3D Cartesian point. Unlike mechanical scanning systems, which accumulate measurements as the sensor rotates, compact multibeam systems can produce volumetric observations in real time from a moving platform. These observations provide explicit scene geometry under poor optical visibility, while RGB cameras provide appearance and semantic information that is difficult to recover from acoustic measurements alone. Pairing the two modalities therefore offers a complementary representation of underwater scenes.

\begin{figure*}[t]
\centering
\includegraphics[width=\textwidth]{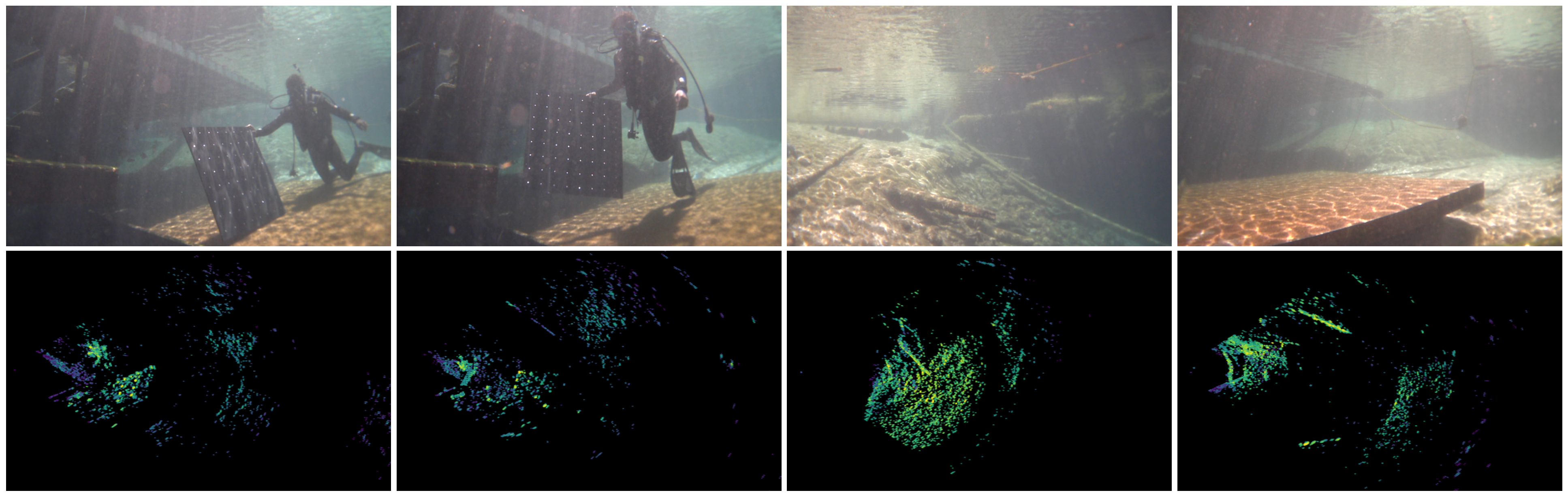}
\vspace{-3mm}
\caption{Representative synchronized observations from uScenes. The top row shows RGB images, and the bottom row shows bird's eye view projections of the corresponding 3D sonar point clouds.}
\label{fig:overview}
\vspace{-3mm}
\end{figure*}

Existing underwater datasets do not fully support this combination. Multimodal datasets such as RGBS50\cite{Li2025RGBS50}, HODOR\cite{Wilts2025HODOR}, and R-S9 pair optical observations with forward looking sonar imagery~\cite{Zhang2026SVFNet}. In contrast, BrSPCD\cite{Huang2025BrSPCD} and SUOP\cite{Ha2026SUOP} provide sonar derived 3D point clouds without synchronized RGB imagery. MAOUD \cite{Chu2026MAOUD} contains both optical and acoustic data, but its 3D geometry is obtained from a laser scanner rather than directly from the sonar.

To address this gap, we introduce \textbf{uScenes}, a multimodal underwater dataset containing synchronized RGB images and 3D multibeam sonar point clouds. The data were collected from a mobile BlueROV2 at Blue Grotto, Florida, across multiple sessions, as illustrated in Fig. \ref{fig:illustrative_example}. The dataset contains 110 scenes and 95,834 synchronized observation pairs, representing 277.6 minutes of underwater recordings. The scenes contain divers, fish, underwater robots, submerged structures, pipes, platforms, and cave passages observed under varied trajectories and visibility conditions.

Each synchronized observation associates RGB appearance with acoustic returns expressed directly in the sonar coordinate frame, as shown in Fig. \ref{fig:overview}. The released data include RGB images, metric 3D sonar points, normalized signal strength, timestamps, scene metadata, and explicit coordinate conventions. The resulting dataset provides a foundation for underwater sensor fusion, cross modal representation learning, 3D scene understanding, and perception under degraded optical conditions\cite{grover2026embodied, dong2026post}.


The main contributions of this work are summarized as follows:
\begin{itemize}
\item We introduce uScenes, a field collected underwater dataset containing 110 scenes, 95,834 synchronized RGB and 3D multibeam sonar observations, and 277.6 minutes of recordings acquired from a mobile robotic platform.
\item We provide metric sonar point clouds containing 3D position and normalized signal strength, together with paired RGB images, timestamps, scene metadata, and clearly documented coordinate conventions.
\item We describe the complete dataset construction methodology, including the sensing platform, temporal association, point cloud generation, data organization, and optic acoustic calibration procedures.
\end{itemize}

\section{RELATED WORK}
\label{section:related_work}
















\begin{table*}[!t]
  \centering
  \footnotesize
  \setlength{\tabcolsep}{5pt}
  \renewcommand{\arraystretch}{1.15}
  \caption{\textbf{Comparison with representative underwater perception datasets.}
}
  \begin{tabular}{@{}l|c|c|c|c|c@{}}
    \toprule
    \multicolumn{1}{c}{Dataset}
    & Optical data
    & Sonar output
    & 3D geometry
    & Time-synchronized
    & Collection setting \\
    \midrule
    USIS16K~\cite{hong2025usis16k}
    & RGB images
    & None
    & None
    & N/A\textsuperscript{$\dagger$}
    & Diverse sources \\
    UATD~\cite{xie2022uatd}
    & None
    & 2D images
    & None
    & N/A\textsuperscript{$\dagger$}
    & Lake and shallow water \\
    Marine Debris~\cite{valdenegrotoro2025marinedebris}
    & None
    & 2D images
    & None
    & N/A\textsuperscript{$\dagger$}
    & Water tank and quarry \\
    RGBS50~\cite{Li2025RGBS50}
    & RGB video
    & 2D video
    & None
    & Yes
    & Deep water pool \\
    HODOR~\cite{Wilts2025HODOR}
    & Stereo video
    & Sonar video
    & None
    & Yes
    & Fixed field station \\
    R-S9~\cite{Zhang2026SVFNet}
    & RGB images
    & 2D images
    & None
    & Yes
    & Field environment \\
    MAOUD~\cite{Chu2026MAOUD}
    & RGB images
    & 2D images and video
    & Laser point cloud
    & Yes
    & Simulated environment \\
    SUOP~\cite{Ha2026SUOP}
    & None
    & 3D scans
    & Sonar point cloud
    & N/A\textsuperscript{$\dagger$}
    & Stationary seafloor \\
    \midrule
    \textbf{uScenes}
    & \textbf{RGB images}
    & \textbf{3D returns}
    & \textbf{Sonar point cloud}
    & \textbf{YES}
    & \textbf{Mobile field robot} \\
    \bottomrule
  \end{tabular}
  
  \footnotesize{\textsuperscript{$\dagger$}Single sensing modality; cross-modal synchronization is not applicable.}
  \label{tab:dataset_comparison}
\end{table*}

\subsection{Underwater Perception Modalities}

Underwater perception systems commonly combine optical, acoustic, and navigation sensors because each modality captures different properties of the environment~\cite{Huy2023UnderwaterPerception,Cong2021UnderwaterSensing}. Optical cameras provide color, texture, and detailed appearance information, but image quality deteriorates because of wavelength dependent attenuation, scattering, turbidity, and insufficient illumination~\cite{Hao2024UnderwaterOptical}. Acoustic sensors depend on sound propagation rather than visible light and can observe structures when optical visibility becomes poor~\cite{negahdaripour2007epipolar}. However, acoustic images contain speckle, weak texture, and sensor dependent distortions that complicate the direct application of conventional visual feature detectors~\cite{kyatham2024sonarfeatures}.

Most imaging sonars represent acoustic measurements in a two dimensional range and bearing image. Because elevation is not directly resolved, a return within a single sonar frame can correspond to multiple positions in 3D space. Negahdaripour estimated 3D sensor motion from feature tracks observed across two dimensional forward looking sonar video~\cite{negahdaripour2013motion}. Huang and Kaess further demonstrated that acoustic structure from motion can recover scene geometry from multiple sonar observations~\cite{Huang2015ASFM}. More recent work reconstructs underwater objects from multiple forward looking sonar views while modeling multipath reflections near the water surface~\cite{liu2024objectmodeling}. These approaches demonstrate that two dimensional sonar does not prevent 3D reconstruction, but reconstruction requires platform motion, reliable correspondences, and additional geometric constraints. In contrast, 3D multibeam sonar measures both horizontal and vertical angles within each scan, allowing valid range observations to be represented directly as metric 3D points~\cite{burgul2025compact3dsonar}.

\subsection{Optical and Acoustic Datasets}

Public underwater optical datasets have supported image enhancement, object detection, and semantic or instance segmentation. USIS16K contains 16,151 underwater images with annotations spanning 158 object categories and provides benchmarks for salient instance segmentation and object detection~\cite{hong2025usis16k}. Such datasets capture substantial appearance diversity but do not contain acoustic measurements that remain available when optical visibility deteriorates.

Acoustic datasets provide an alternative representation for underwater perception. UATD contains more than 9,000 multibeam forward looking sonar images with annotations for ten target categories collected in lake and shallow water environments~\cite{xie2022uatd}. The Marine Debris Forward Looking Sonar datasets extend this setting across water tank, turntable, and flooded quarry environments and support classification, detection, segmentation, and feature learning~\cite{valdenegrotoro2025marinedebris}. These datasets have advanced learning from sonar imagery, although their acoustic observations remain two dimensional and are not paired with simultaneous optical measurements as.

\subsection{Multimodal Optic Acoustic Datasets}

Optic acoustic fusion has been studied as a geometric sensing problem before the development of recent multimodal datasets. Negahdaripour derived the epipolar geometry relating optical and acoustic image measurements and showed how the different sensor projection models constrain cross modal correspondences~\cite{negahdaripour2007epipolar}. Negahdaripour et al. subsequently introduced joint system calibration and 3D reconstruction methods based on corresponding optical and sonar measurements~\cite{negahdaripour2009optiacoustic}. More recent work extended this framework to GoPro cameras and forward looking sonar, using ray tracing to model refraction through the camera housing and estimate the relative sensor poses~\cite{negahdaripour2024gopro}. These studies establish the geometric foundation for optic acoustic fusion and emphasize that synchronized observations require an appropriate calibration model before precise spatial association can be performed.

Recent datasets have begun to provide synchronized optical and acoustic observations for learning based perception. RGBS50 contains 50 temporally aligned RGB and sonar video sequences with more than 87,000 bounding box annotations for underwater object tracking~\cite{Li2025RGBS50}. The R-S9 dataset provides 3,732 aligned RGB and sonar image pairs for underwater fusion classification~\cite{Zhang2026SVFNet}. Both datasets demonstrate the value of learning complementary features across optical and acoustic images.

HODOR offers over 400 hours of synchronized stereo camera and sonar video from a stationary underwater fish observatory in the Kiel Fjord~\cite{Wilts2025HODOR}, making it valuable for fish monitoring, biomass estimation, and long-term behavioral analysis. Its stationary platform and video-based acoustic representation, however, target a different setting from mobile robot perception with metric 3D acoustic measurements.

Simulation provides another mechanism for generating multimodal underwater data under controlled conditions. HoloOcean and OceanSim support repeatable marine robotics experiments and configurable sensor simulation~\cite{potokar2022holocean,oceansim2025}. MAOUD~\cite{Chu2026MAOUD} combines RGB images, acoustic images and videos, and laser point clouds collected in a controlled underwater simulation environment. Its laser point clouds provide geometric information, while its acoustic modality is represented through images and videos. Real field recordings remain complementary because they capture platform motion, natural visibility changes, acoustic reflections, and environmental structures encountered during deployment, as summarized in Table~\ref{tab:dataset_comparison}.

\subsection{3D Sonar Perception}

Three dimensional sonar point clouds have recently been investigated for inspection, object recognition, and mapping. BrSPCD provides labeled sonar point clouds of underwater bridge structures for semantic segmentation~\cite{Huang2025BrSPCD}. The SUOP dataset contains 1,555 mechanical scanning sonar point clouds of five small underwater object categories, together with raw sonar scans, metadata, and corresponding two dimensional sonar images~\cite{Ha2026SUOP}. These datasets demonstrate the value of acoustic point clouds for recovering object shape and structural geometry, but they focus on stationary scanning and do not include synchronized RGB observations from a mobile robot.

Burgul et al. mounted a compact 3D multibeam sonar on a BlueROV2 and demonstrated its use for scan registration, localization, and dense underwater mapping~\cite{burgul2025compact3dsonar}. Their work also introduced an optic acoustic calibration procedure based on corresponding cinder block corners observed by the camera and sonar. The resulting 3D and 2D correspondences were formulated as a perspective (n) point problem and solved using EPnP~\cite{lepetit2009epnp}. This work establishes the sensing and geometric potential of compact 3D sonar, with its primary emphasis placed on calibration, state estimation, and mapping.

uScenes complements these efforts by providing a large, scene organized collection of synchronized RGB images and compact 3D sonar point clouds acquired from a mobile underwater robot. Its distinguishing characteristic is not simply the presence of optical and acoustic sensors, but that the acoustic modality itself provides metric 3D returns for every synchronized observation. This representation supports research connecting visual appearance with acoustic geometry across dynamic field recordings.
\section{DATASET}
\label{sec:methodology}

\subsection{Sensor Platform and Data Collection}
\label{sec}

\begin{figure}[t]
\centering
\includegraphics[width=0.9\columnwidth]{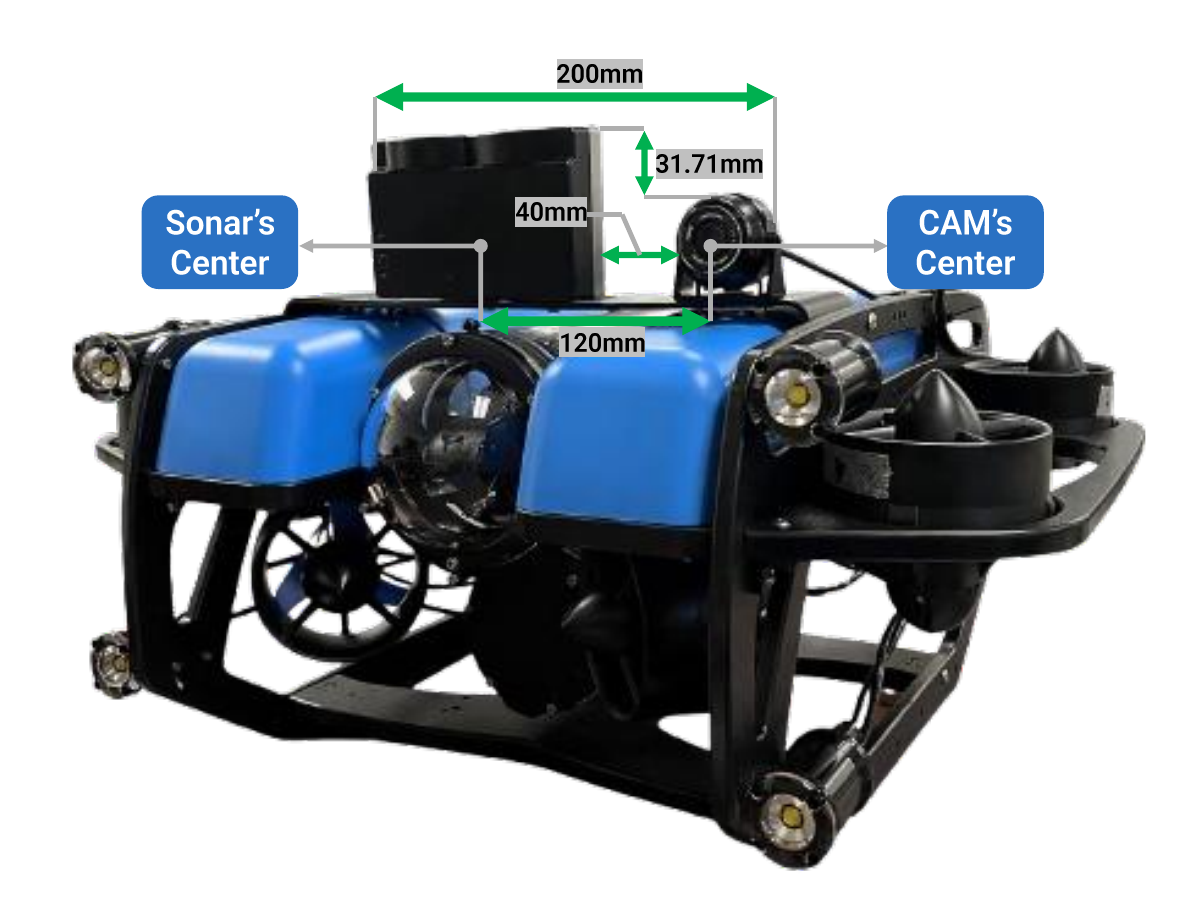}
\caption{The uScenes data collection platform. A BlueROV2 carries an RGB camera and a 3D multibeam sonar in a rigid, forward facing configuration.}
\label{fig}
\end{figure}

The uScenes sensing platform consists of a BlueROV2 equipped with a DWE underwater RGB camera and a Water Linked Sonar 3D-15 \cite{waterlinked_sonar3d15}. The sensor arrangement is shown in Fig.~\ref{fig}, and the specifications of both sensors are summarized in Table~\ref{tab:sensors}. The sonar was mounted near the center of the vehicle, while the camera was positioned laterally with an overlapping forward field of view. The sensor centers were separated by 120 mm laterally and 31.71 mm vertically, corresponding to a total displacement of approximately 124 mm. The rigid mounting configuration remained unchanged throughout data collection.

The RGB camera records ($1280 \times 720$) images at approximately 30 Hz, while the sonar produces ($256 \times 64$) range observations at approximately 6 Hz. Each valid sonar measurement provides range and registered signal strength, from which a 3D point cloud is constructed. Depending on the observed scene, a sonar frame contains between 1 and 12,627 returns, with an average of 6,805 returns.

\begin{table}[t]
\centering
\caption{Sensor configuration.}
\label{tab:sensors}
\small
\setlength{\tabcolsep}{3.4pt}
\begin{tabular}{p{1.25cm}p{2.05cm}p{3.75cm}}
\toprule
Sensor & Property & Value \\
\midrule
Camera & Resolution & $1280\times720$ pixels \\
       & Rate & Approximately 30 Hz \\
       & Model & Pinhole with five radial and tangential distortion terms \\
       & Intrinsics & $f_x=922.26$, $f_y=920.86$, $c_x=694.06$, $c_y=387.33$ pixels \\
       & Radial & $k_1=-0.3496$, $k_2=0.3192$ \\
       & & $k_3=-0.2017$ \\
       & Tangential & $p_1=-0.0045$ \\
       & & $p_2=-0.0051$ \\
\midrule
3D sonar & Range image & $256\times64$ samples \\
         & Field of view & $90^{\circ}$ horizontal, $40^{\circ}$ vertical \\
         & Rate & Approximately 6 Hz \\
         & Output & Range and registered signal strength \\
         & Returns & 1 to 12,627 points per frame, mean 6,805 \\
\bottomrule
\end{tabular}
\end{table}

Data were collected at Blue Grotto in Williston, Florida, during seven field sessions. The vehicle recorded natural and human influenced underwater environments containing divers, fish, underwater robots, submerged platforms, pipes, and cave passages. Dedicated sequences containing calibration targets were also collected. The recordings were divided into 110 scenes representing 39 collection tasks, resulting in 95,834 synchronized observations and 277.6 minutes of data.

\subsection{Temporal Association}

The camera and sonar operate at different acquisition rates and produce independent timestamps. Each sonar observation is associated with the camera image having the nearest timestamp. Given a sonar timestamp ($t_s$) and camera timestamps ($t_{c,i}$), the selected camera observation is

\begin{equation}
i^{*} = \arg\min_i \left|t_{c,i} - t_s\right|,
\qquad
\left|t_{c,i^{*}} - t_s\right| \leq 83~\mathrm{ms}.
\label{eq:temporal_association}
\end{equation}

The 83 ms threshold corresponds to approximately half of one sonar acquisition interval. Across the accepted observation pairs, the mean absolute timestamp difference is 13.2 ms, the median is 11.2 ms, and 95 percent of the pairs differ by no more than 32.8 ms. The sonar timestamp represents the midpoint of its sequential scan. The association therefore aligns each camera image with the center of the sonar acquisition period, although motion occurring within a sonar scan remains present.

\begin{figure*}[t]
\centering

\includegraphics[
    width=\textwidth,
    trim=0 70mm 0 30mm,
    clip
]{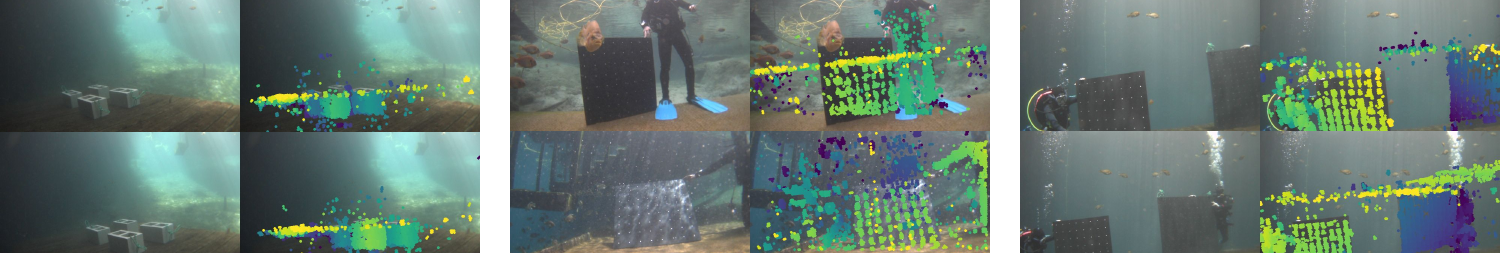}

\makebox[0.333\textwidth][c]{\textbf{(a)}}%
\makebox[0.333\textwidth][c]{\textbf{(b)}}%
\makebox[0.333\textwidth][c]{\textbf{(c)}}

\caption{Calibration target configurations used during uScenes data collection. (a) Four dimension constrained concrete blocks placed on the submerged platform. (b) A diver holding a calibration board containing round headed pins. (c) Two calibration boards containing flat headed screws and positioned at different depths. Within each group, the RGB observation is shown alongside the corresponding 3D sonar returns used for selecting calibration points.}
\label{fig:calibration_targets}

\end{figure*}

\subsection{Sonar Point Cloud Construction}

Let ((u,v)) represent a location in the sonar range image, with width (W), height (H), horizontal field of view ($\phi_h$), and vertical field of view ($\phi_v$). The corresponding horizontal and vertical angles are

\begin{equation}
\theta =
\frac{u}{W-1}\phi_h-\frac{\phi_h}{2},
\qquad
\psi =
\frac{v}{H-1}\phi_v-\frac{\phi_v}{2}.
\label{eq:sonar_angles}
\end{equation}

For a measured range (r), the Cartesian return in the vendor sonar coordinate frame is computed as

\begin{equation}
\mathbf{p}^{s} =
r
\begin{bmatrix}
\cos\psi\cos\theta \
\cos\psi\sin\theta \
\sin\psi
\end{bmatrix}.
\label{eq:sonar_point}
\end{equation}

This coordinate frame is right handed, with (x) pointing forward, (y) pointing right, and (z) pointing downward. The processed binary files use the opposite lateral sign. Consequently, points loaded from the dataset are converted to the vendor frame using

\begin{equation}
    \mathbf{p}^{s} = \operatorname{diag}(1,-1,1)\mathbf{p}^{\mathrm{bin}}
    \label{eq:coordinate_conversion}
\end{equation}

before applying a rigid sensor transformation.

The acoustic signal (a) is compensated for range using ($\widetilde{s}=ar^2$). It is then normalized by the 99th percentile within the current sonar observation and clipped to the interval ([0,1]). The normalized value represents relative signal strength within a frame rather than calibrated acoustic reflectivity.

\subsection{Dataset Organization}

\begin{table}[t]
\centering
\caption{Scene level summary by primary object of interest.}
\label{tab:scene_distribution}
\footnotesize
\setlength{\tabcolsep}{3pt}
\renewcommand{\arraystretch}{1.1}

\begin{tabular}{@{}lrrrrr@{}}
\toprule
Class & Scenes & \% & Pairs & \% & Minutes \\
\midrule
None      & 46  & 41.8 & 42,957 & 44.8 & 125.7 \\
Human     & 24  & 21.8 & 24,167 & 25.2 & 69.6  \\
Structure & 30  & 27.3 & 20,979 & 21.9 & 60.5  \\
ROV       & 4   & 3.6  & 4,312  & 4.5  & 12.1  \\
Fish      & 6   & 5.5  & 3,419  & 3.6  & 9.7   \\
\midrule
\textbf{Total}
& \textbf{110}
& \textbf{100.0}
& \textbf{95,834}
& \textbf{100.0}
& \textbf{277.6} \\
\bottomrule
\end{tabular}
\end{table}

Each scene contains the paired camera images, sonar point clouds, and timestamp metadata. Camera observations are stored as JPEG images. Each sonar observation is stored as a binary floating point array with shape $(N \times 4)$, where each row contains ([x,y,z,s]). The fourth value is the normalized acoustic signal strength.

A JSON Lines manifest records the camera timestamp, sonar timestamp, temporal difference, frame identifier, and number of sonar returns for every pair. A global scene index records session information, duration, and scene level task descriptions. These descriptions provide coarse information about the primary content of a recording and are not intended as frame level detection annotations.

Table~\ref{tab:scene_distribution} summarizes the dataset according to the primary object of interest assigned to each scene. Human and structure recordings account for 54 of the 110 scenes, while 46 scenes contain no designated object of interest and instead capture general underwater environments. These categories provide a coarse index for selecting recordings rather than object annotations.

\subsection{Optic Acoustic Calibration Methodology}
\label{sec}

Optic acoustic calibration establishes the spatial relationship between the camera and sonar. Temporal synchronization identifies observations acquired at approximately the same time, but it does not determine where a 3D sonar return appears in the camera image. This association requires the camera intrinsics and a rigid transformation from the sonar coordinate frame to the camera coordinate frame.

Camera intrinsics were estimated underwater using multiple observations of an ($8 \times 8$) white pin circle grid. A pinhole camera model with radial and tangential distortion was fitted to the detected grid points. Performing this procedure underwater allows the effective projection model to account for the dominant imaging effects introduced by the camera housing and water interface.

We consider three target configurations for estimating the optic acoustic extrinsic transformation, as illustrated in Fig.~\ref{fig:calibration_targets}. Each configuration provides identifiable 3D points in the sonar coordinate frame and their corresponding 2D locations in the camera image.

The first configuration follows the concrete block target \cite{burgul2025compact3dsonar}. Four concrete masonry blocks with known dimensions are placed within the overlapping fields of view of the sensors. In each RGB image, the visible corners of the upper block surfaces are selected as 2D image points. In the sonar point cloud, returns belonging to each upper surface are isolated and fitted with a rectangle constrained by the known block dimensions. The corners of the fitted rectangle provide the corresponding 3D sonar points. A consistent ordering of the corners is maintained for every block and observation.

The second configuration follows the round headed pin board \cite{negahdaripour2024gopro}. The board contains an ($8 \times 8$) grid of pins with known spacing and is held at different positions and orientations within the common sensor view. The center of each visible pin is selected in the RGB image. Corresponding acoustic peaks are then identified in the 3D sonar point cloud. The known row and column arrangement of the grid determines the correspondence between each image point and sonar return.

The third configuration uses calibration boards containing flat headed screws. We observed that the flat screw heads produce more isolated acoustic responses than the round headed pins by reducing responses across neighboring sonar beams. This makes individual reflectors easier to identify in the 3D point cloud. Two boards are positioned at different depths to provide correspondences across a larger spatial volume. Screw centers are selected in the RGB images, while the corresponding compact acoustic returns are selected from the sonar point clouds. Board identity and grid position are used to maintain consistent correspondence labels.

For all three configurations, let $\mathbf{p}^{s}_{j}$ denote a selected 3D sonar point and let $\mathbf{q}_{j}$ denote its corresponding 2D image location. The rigid transformation from the sonar coordinate frame to the camera coordinate frame is

\begin{equation}
\mathbf{p}^{c}_{j} = \mathbf{R}_{cs}\,\mathbf{p}^{s}_{j} + \mathbf{t}_{cs},
\label{eq:sonar_to_cam}
\end{equation}

where $\mathbf{R}_{cs} \in SO(3)$ is the rotation matrix and $\mathbf{t}_{cs} \in \mathbb{R}^{3}$ is the translation vector. The extrinsic parameters are estimated by minimizing the reprojection error over the collected correspondences:

\begin{equation}
\min_{\mathbf{R}_{cs},\,\mathbf{t}_{cs}} \sum_{j} \left\| \pi\!
\left(\mathbf{R}_{cs}\,\mathbf{p}^{s}_{j} + \mathbf{t}_{cs}\right) - \mathbf{q}_{j} \right\|_{2}^{2}.
\label{eq:reproj_error}
\end{equation}

where ($\pi(\cdot)$) denotes the underwater camera projection and distortion model, and ($\rho(\cdot)$) is a robust penalty. EPnP~\cite{lepetit2009epnp} is used to initialize the rotation and translation, followed by nonlinear refinement over all selected correspondences. Because the camera and sonar are rigidly mounted, all target configurations and observations share a single rotation matrix and translation vector.
\section{DISCUSSION}
\label{sec}

\subsection{Research Opportunities}

uScenes addresses a gap between underwater optical datasets, which provide appearance information without measured 3D geometry, and conventional sonar datasets, which primarily contain two dimensional acoustic images. The synchronized observations provide complementary descriptions of the same underwater scenes. RGB images capture color, texture, and object appearance, while the sonar point clouds provide range and three dimensional structure under poor optical conditions.

This combination supports research in cross modal representation learning, sonar assisted object perception, and multimodal 3D scene understanding. Temporal synchronization also allows models to learn relationships between the modalities without requiring direct point to pixel correspondence. With additional annotations, the dataset can support future benchmarks for underwater object detection, segmentation, and semantic understanding using optical and acoustic observations.

\subsection{Calibration Considerations}

Precise spatial association between the modalities requires an accurate transformation between the sonar and camera coordinate frames. Section~\ref{sec} describes the calibration targets, correspondence construction, and optimization procedure considered for the sensing platform. 

The current dataset should therefore be treated as temporally synchronized but not spatially registered. The paired observations can be used for tasks based on temporal correspondence, scene level association, or independent modality processing. Tasks that require projecting individual sonar points into the camera image should first establish and validate an appropriate optic acoustic calibration. The included calibration sequences provide material for developing and comparing such methods.

\subsection{Limitations}

The current dataset was collected using one robotic platform at a single freshwater site. Although the recordings contain varied objects, structures, visibility conditions, and vehicle trajectories, they do not establish generalization across different sensors, platforms, geographic locations, or saltwater environments.

Dense object annotations are not included in the current release. Scene descriptions indicate the primary content of each recording but should not be interpreted as frame level ground truth. Additional 2D and 3D annotations will be required to establish supervised detection and segmentation benchmarks.

The released acoustic signal strength is compensated for range and normalized independently within each sonar frame. It represents relative signal strength rather than calibrated acoustic reflectivity and should not be directly compared across frames. In addition, each sonar observation is acquired sequentially and timestamped at the midpoint of the scan. Motion of the vehicle or observed objects during acquisition can therefore distort the reconstructed point cloud. Finally, the underwater camera calibration uses an effective pinhole model, while the physical effect of the flat port housing has not been fully validated.
\section{Conclusions and Future Work}
\label{sec:conclusion}

We introduced uScenes, a multimodal underwater dataset containing synchronized RGB images and 3D multibeam sonar point clouds collected from a mobile field robot. The dataset contains 95,834 paired observations across 110 scenes, representing 277.6 minutes of recordings from seven field sessions. We described the sensing platform, data collection process, temporal association, sonar point cloud construction, dataset organization, and the calibration methods considered for the platform.

By pairing optical appearance with directly measured acoustic geometry, uScenes provides a foundation for underwater sensor fusion, cross modal representation learning, and 3D scene understanding. Future work will focus on establishing a reliable optic acoustic calibration, expanding the annotations, and collecting data across additional platforms and underwater environments.

\section*{Acknowledgment}
The authors would like to thank Andres Pulido, Nikhil Iyer, Cayman Christ, and Cheryl Thacker for their help with diving, and Blue Grotto Dive Resort for the facilities for data collection. This work was supported by New Faculty Startup Fund from Seoul National University, CRADA22-0034-J003, and NVIDIA Academic Grant Program.

\bibliographystyle{IEEEtran}
\bibliography{root}

\end{document}